\documentclass{article}

\usepackage{PRIMEarxiv}

\usepackage[utf8]{inputenc} 
\usepackage[T1]{fontenc}    
\usepackage{hyperref}       
\usepackage{url}            
\usepackage{booktabs}       
\usepackage{amsfonts}       
\usepackage{nicefrac}       
\usepackage{microtype}      
\usepackage{lipsum}
\usepackage{fancyhdr}       
\usepackage{graphicx}       
\graphicspath{{media/}}     
\usepackage{amsmath}
\usepackage{hanging}
\usepackage{listings}
\usepackage{color}
\usepackage{ccaption}

\definecolor{dkgreen}{rgb}{0,0.6,0}
\definecolor{gray}{rgb}{0.5,0.5,0.5}
\definecolor{mauve}{rgb}{0.58,0,0.82}

\title{Two-step hyperparameter optimization method: Accelerating hyperparameter search by using a fraction of a training dataset
\thanks{This work has been submitted to Artificial Intelligence for the Earth Systems (ISSN: 2769-7525). Copyright in this work may be transferred without further notice.}
}

\author{
Sungduk Yu\\
  Department of Earth System Science \\
  University of California Irvine \\
  Irvine, CA\\
  \texttt{sungduk@uci.edu} \\
  \And
  Mike Pritchard\\
  Department of Earth System Science \\
  University of California Irvine \\
  Irvine, CA\\
   \And
  Po-Lun Ma\\
  Pacific Northwest National Laboratory \\
  Richland, WA\\
  \AND
  Balwinder Singh\\
  Pacific Northwest National Laboratory \\
  Richland, WA\\
  \And
  Sam Silva\\
  Department of Earth Sciences \\
  University of Southern California \\
  Los Angeles, CA \\
}

\begin{document}
\maketitle

\begin{abstract}
Hyperparameter optimization (HPO) is an important step in machine learning (ML) model development, but common practices are archaic—primarily relying on manual or grid searches. This is partly because adopting advanced HPO algorithms introduces added complexity to the workflow, leading to longer computation times. This poses a notable challenge to ML applications, as suboptimal hyperparameter selections curtail the potential of ML model performance, ultimately obstructing the full exploitation of ML techniques. In this article, we present a two-step HPO method as a strategic solution to curbing computational demands and wait times, gleaned from practical experiences in applied ML parameterization work. The initial phase involves a preliminary evaluation of hyperparameters on a small subset of the training dataset, followed by a re-evaluation of the top-performing candidate models post-retraining with the entire training dataset. This two-step HPO method is universally applicable across HPO search algorithms, and we argue it has attractive efficiency gains.

As a case study, we present our recent application of the two-step HPO method to the development of neural network emulators for aerosol activation. Although our primary use case is a data-rich limit with many millions of samples, we also find that using up to 0.0025\% of the data—a few thousand samples—in the initial step is sufficient to find optimal hyperparameter configurations from much more extensive sampling, achieving up to 135$\times$ speed-up. The benefits of this method materialize through an assessment of hyperparameters and model performance, revealing the minimal model complexity required to achieve the best performance. The assortment of top-performing models harvested from the HPO process allows us to choose a high-performing model with a low inference cost for efficient use in global climate models (GCMs).
\end{abstract}


\section{Introduction}
The application of artificial intelligence/machine learning (AI/ML) techniques in earth system science is becoming increasingly popular in recent years. However, our field’s adaptation of modern hyperparameter optimization (HPO) or tuning techniques—i.e., advanced search algorithms and extensive search space—has been slow. For example, a manual or a grid search is still commonly used in published studies despite the availability of ample HPO software packages and GPU resources. The question naturally arises: do our ML applications exploit the full potential of ML models? This issue becomes more pressing as ML architectures are getting more sophisticated, i.e., more hyperparameters need to be optimized, “the curse of dimensionality”.

We suspect that the perceived large investment of time and compute is a major barrier preventing the widespread adoption of modern HPO practices. There are certainly modern ways to reduce computation time: for example, distributed search (i.e., evaluating multiple hyperparameter configurations simultaneously), adaptive search algorithms (i.e., selecting hyperparameter configurations based on the result of preceding hyperparameter evaluations; e.g., Bayesian methods (Snoek et al. 2012)), and adaptive resource allocations (i.e., allocating more resources to promising hyperparameter configurations using a tournament selection, e.g., Hyperband (Li et al. 2016)). Nonetheless, an HPO task can still take a significant amount of time, as both training dataset sizes and ML model complexity have been increasing.

Beyond the conventional strategies such as parallel computing and adaptive algorithms, we further explore accelerating HPO tasks based on a core-set (also known as instance selection) approach, which focuses on reducing the size of the dataset by selecting samples representative of the original dataset. The core-set approach has gained much attention in the past decade in the computer science domain to shorten model training and data processing time for big data (Feldman 2020; Mirzasoleiman et al. 2020; Killamsetty et al. 2021a,b, 2022). Specifically for HPO, Wendt et al. (2020) presented an interesting 2-phase HPO approach: using a small subset for a wide search with low granularity and then using a full dataset for a final search within the narrowed domain identified from the wide search. They showed a 10\% subset combined with a random search algorithm can reliably return the same results as a single-phase random search but 7 times faster. Furthermore, Visalpara et al. (2021) showed that HPO with a 5\% subset can achieve a result comparable to HPO with a full dataset. Despite these promising reports, the applicability of a core-set approach still needs verification for a real-world problem since these studies are based on toy datasets that contain only several 10,000 instances (e.g., CIFAR-100 and MNIST). Testing comparable ideas in the context of an applied machine-learning parameterization problem relevant to hybrid AI climate simulation is new.

In this article, we test a core-set approach for a real-world earth system modeling problem and propose a \textit{two-step HPO method}, which could be viewed as a variant of Wendt et al. (2020) and Li et al. (2016) but with much simpler applicability. Our method is designed to be a general framework that is independent of the specific search algorithms or computing environments used. The central idea behind our method is that HPO applied initially to a small subset of a training dataset can effectively identify optimal hyperparameter configurations, thus reducing the overall computational cost. We demonstrate the effectiveness of our approach through a case study in which we apply our two-step HPO method to optimize a neural network emulator for aerosol activation processes. Our results show that the two-step HPO method can effectively discover optimal hyperparameter configurations using only a small subset of the training dataset (as low as 0.025\% in our case).

The rest of the article is organized as follows. In Section 2, we present our two-step HPO method, including an estimation of cost savings. Section 3 is a case study of our method applied to optimize a neural network emulator. Section 4 covers the computational setup of our case study. Results are presented in Section 5, and Section 6 concludes with a summary of findings and potential implications.

\section{Computational cost saving by the two-step HPO method}
Our two-step HPO method is designed to reduce the computational burden of traditional HPO by using a small subset of a training dataset. In the first step, a large number of trials are conducted using only a small portion of a training dataset. In the second step, the top candidates shortlisted from the first step are retrained with a full training dataset for the final selection (Figure 1a). The second step also serves the purpose of recalibrating model weights with a full training dataset, resulting in final models that are ready for deployment. This approach assumes that hyperparameter evaluations using a subset are indicative of using the entire dataset in data-rich limits. A model trained with a small dataset is expected to be less accurate than one trained with a large dataset. However, if this assumption is valid, the accuracy of top models after retraining in Step 2 should converge regardless of the size of the subset used in Step 1. This will be empirically tested in the case study presented in the following sections.

Our two-step HPO method can meaningfully reduce the computational burden required for HPO, compared to the traditional one-step approach. Two parameters must be preset to use this method: a portion of an entire training dataset to be used in Step 1 ($p_{subset}$) and a portion of total hyperparameter evaluation trials from Step 1 to be retrained in Step 2 ($p_{retrain}$). Then, the fractional computational cost of the two-step approach is approximately equal to $p_{subset} + p_{retrain}$, i.e.,\\
\begin{equation*}
\begin{aligned}
\frac{\textup{Two step HPO}}{\textup{(Traditional) one step HPO}} &= \frac{\textup{Step 1 HPO + Step 2 HPO}}{\textup{(Traditional) one step HPO}}\\
&\approx \frac{p_{subset}\cdot X + (p_{retrain}\cdot n_{trials})\cdot \bar{x}}{X}\\
&=p_{subset}+p_{retrain},
\end{aligned}
\end{equation*}

where X is the total computation cost (e.g., GPU-hour) of a traditional one-step HPO for evaluating $n_{trials}$ number of hyperparameter configurations, and is $\bar{x}$ an average computational cost for evaluating one trial with a full training dataset (i.e., $\bar{x}=X/n_{trials}$). This estimation assumes the training time linearly scales to the size of a training dataset for a given software and hardware environment. 

\begin{figure}
  \centering
  \includegraphics[trim={7.5cm 7.5cm 7.5cm 7.5cm},clip, width=.95\textwidth]{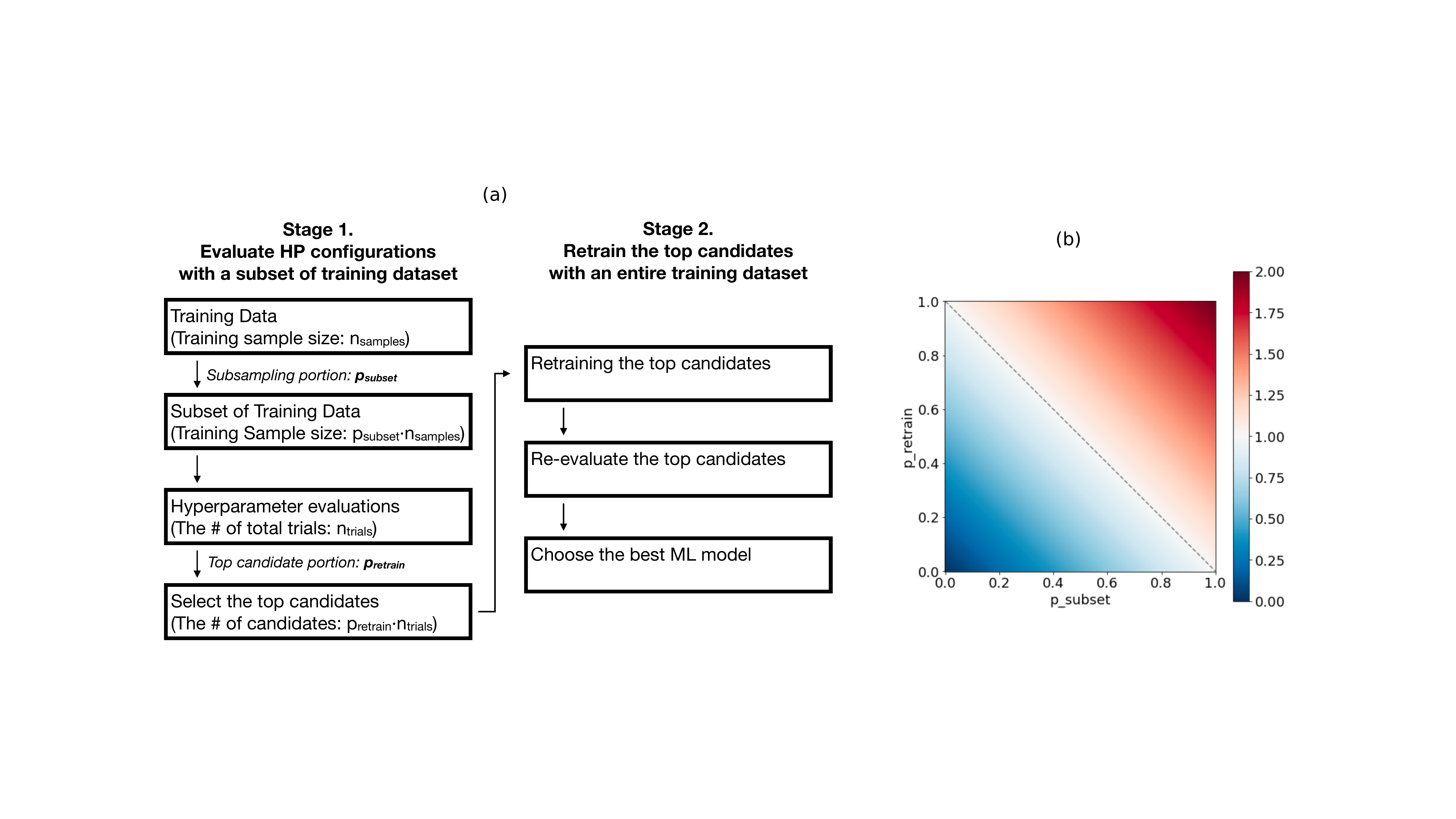}
  \caption{(a) A schematic of the two-step HPO method. In step 1, A $p_{subset}$ portion of an entire training dataset with $n_{samples}$ samples are used for HPO with $n_{trials}$ trials. In step 2, the $p_{retrain}$ portion of $n_{trials}$ trials are retrained with the entire training dataset for final evaluations to select the best hyperparameter configurations. (b) An approximation of the fraction of computational resources (e.g., GPU-hours) required for the two-step HPO method with respect to the traditional one-step approach (i.e., $p_{subset}$ = 1 and $p_{retrain}$ = 0). The savings in computational cost depend on the values of $p_{subset}$ and $p_{retrain}$. As an example, the computational cost reduces to 10\% if 5\% of the dataset is used in step 1 and the top 5\% candidate models are retrained in step 2 (i.e., $p_{subset}$ = 0.05 and $p_{retrain}$ = 0.05). It is worth noting that the traditional approach is a special case of the two-step HPO method with $p_{subset}$ = 1 and $p_{retrain}$ = 0. See Section 2 for more information.}
  \label{fig:fig1}
\end{figure}

\section{Case study: Aerosol activation emulator}
To demonstrate the effectiveness of the two-step HPO, we present our recent HPO work for aerosol activation emulators. Aerosol activation (or “droplet nucleation”) is the process that describes the spontaneous growth of aerosol particles into cloud droplets in an ascending air parcel. This process occurs at 10-100 µm scales and accordingly is parameterized with varying degrees of assumptions in weather and climate models of which the grid resolutions are orders of magnitude larger (Abdul‐Razzak and Ghan 2000; Ghan et al. 2011). Instead of relying on such parameterizations, a neural network-based model enables direct emulation of detailed cloud parcel models which more explicitly simulate the droplet nucleation process in a rising air parcel with minimal assumptions. 

We generate a dataset for developing deep neural network emulators using explicit numerical calculations of aerosol activation based on Ghan et al. (2011) cloud parcel model. The initial conditions for the cloud parcel model simulations are populated by harvesting hourly instantaneous atmospheric state variables from a year-long simulation of U.S. Department of Energy’s Energy Exascale Earth System Model (E3SM) (Golaz et al. 2019) Atmosphere Model (EAM) version 1 (Rasch et al. 2019). A total of 98.6 million training samples are obtained and are divided into a train/validation dataset for HPO (19.7 million samples; 20\%) and a holdout test dataset (78.8 million samples; 80\%). The input and output variables for the emulators are listed below. Note that the aerosol-related variables have a dimension of four (“nmodes”) as both E3SM and the cloud parcel model are set up with four aerosol modes corresponding to the 4-mode version of the modal aerosol module (MAM4) (Liu et al. 2016; Wang et al. 2020).

Inputs:
\begin{itemize}
    \item “Tair”: Temperature, K
    \item “Pressure”: Pressure, hPa
    \item “rh”:  relative humidity
    \item “wbar”: vertical velocity, cm/s
    \item “num\_aer [nmodes]”: aerosol number, 1/cm3
    \item “r\_aer [nmodes]”: aerosol dry radius, µm
    \item “kappa [nmodes]”: hygroscopicity
\end{itemize}
Outputs:
\begin{itemize}
    \item “fn [nmodes]”: activated fraction
\end{itemize}

The input variables are standardized using z-scores, but the output variable is intrinsically bounded by zero and one.

\section{Hyperparameter optimization}
\subsection{Two-step HPO setup}
For good sampling, a computationally ambitious HPO project with an identical setup is repeated four times with decreasing subset sizes for Step 1: 100\%, 50\%, 25\%, 5\%, 0.5\%, 0.05\%, and 0.025\%  (i.e., $p_{subset}$ = 1.00, 0.50, 0.25, 0.05, 0.005, 0.0005, and 0.00025 respectively) containing about 19.7M, 9.9M, 4.9M, 1.0M, 98.6K, 9.9K, and 4.9k samples, respectively. Hereafter, these projects will be referred to as P1.00, P0.50, P0.25, P0.05, P0.005, P0.0005, and P0.00025 projects. The P1.00 is served as a control experiment that the rest subset HPO projects are evaluated against. Each project includes an unusually expansive 10,000 trials of hyperparameter evaluation (i.e., $n_{trials}$ = 10000). The train/validation split ratio is 80:20 for the P1.00, P0.50, P0.25, and P0.05 and 50:50 for the P0.005, P0.0005, and P0.00025. For Step 2, the top 50 trials from each project are retrained with a full training dataset (i.e., $p_{retrain}$ = 0.005). 

\subsection{Hyperparameter search space}
We use a Multi-layer Perceptron (MLP) as a machine learning architecture for our emulator, which has been applied for emulating physical processes in various applications including aerosol activation and other microphysical processes  (Silva et al. 2021; Gettelman et al. 2021; Chiu et al. 2021; Alfonso and Zamora 2021). Silva et al. (2021) employed a modern HPO workflow but only with a limited number of tuning trials ($\sim$400). To ensure the robust sampling of the hyperparameter search space, we focus on only two key hyperparameters (the number of hidden layers and the number of nodes in each layer) that defines the backbone (i.e., complexity) of the MLP neural networks. However, any arbitrary combinations of hyperparameters (e.g., batch size, learning rate, optimizer, regularization, etc.) can be included to a HPO task. The search space for each hyperparameter is:
\begin{itemize}
    \item The number of hidden layers ($N_{Layers}$): [1, 2, 3, 4, 5, 6, 7, 8, 9, 10 ,  11, 12, 13, 14, 15, 16, 17, 18, 19, 20]; and
	\item The number of nodes ($N_{Nodes}$) in each layer: [8, 16, 32, 64, 128, 256, 512, 1024, 2048].
 \end{itemize}
 Note that $N_{Nodes}$ is selected independently for each hidden layer. That is, for $N_{Layers}=k$, $N_{Nodes}$ is drawn from the search space k times. The rest of the hyperparameters are fixed as follows We use a ReLU activation function for the hidden layers; a sigmoid activation for the output layer; MSE as a loss function; and Adam optimizer (Kingma and Ba 2014) as the gradient descent algorithm with a batch size of 1024 and a learning rate of 0.001. We enforce an early stopping rule with a patience of 5 epochs and maximum training epochs of 100. The neural network weights and validation MSE are recorded when early stopping is called.

\subsection{Search algorithm}
We use a random search algorithm, which is easy to parallelize and fault-tolerant by design. Despite its simplicity, a random search algorithm is known to be more efficient than—or at least as good as—a grid search (Bergstra and Bengio 2012). We note that our two-step HPO can be applied to any HPO algorithm since it only concerns the size of a training dataset. However, our two-step HPO approach may not be suitable for adaptive search algorithms, which rely on the result of preceding hyperparameter evaluations to assign hyperparameter configurations for succeeding evaluations. Unlike a random (or non-adaptive) algorithm that uniformly samples hyperparameters under a given distribution regardless of a training dataset size, an adaptive algorithm may converge towards different optima depending on the training dataset (i.e., the information content of a training dataset). 

\subsection{Computational setup for distributed search}
We use a distributed search setup to speed up the process, using GPU nodes on the Bridges-2 supercomputer at the Pittsburgh Supercomputing Center (Towns et al. 2014). For that, we choose KerasTuner for its seamless integration with Tensorflow/Keras libraries (https://github.com/keras-team/keras-tuner); however, our two-step method should be easily implemented in other popular HPO software frameworks (e.g., Hyperopt, Optuna, and Ray Tune). The distributed mode of KerasTuner uses a manager-worker model. Despite the user-friendly interface of KerasTuner, setting up a distributed search mode on an HPC with a job scheduler can be complex due to dynamic computing environments. For readers’ reference, code examples of how we set up distributed HPO on Bridge-2 are included in the Supplemental Material.

We use two GPU nodes, totaling 16 GPUs (8 GPUs/node; NVIDIA Tesla V100-32GB SXM2). One GPU is assigned for one worker, allowing for concurrent evaluations of 16 hyperparameter configurations. The elapsed wall-clock time for computation scales linearly with the size of the training dataset, e.g., 5.2, 23.4, 48.6, and 94.3 hours for P0.05, P0.25, P0.50, and P1.00 projects, respectively. On the other hand, the HPO projects with smaller subsets, e.g., P0.00025, P0.0005, and P0.005, are conducted with only a half GPU node (4 GPUs) take 2.8, 3.0, and 5.0 wall-clock hours, respectively. Note that the computational resources (e.g., GPU-hour) for P1.00 is about 135 times more than that for P0.00025. The absence of linear scaling in P0.00025–P0.005 is due to the increasing relative portion of computational overhead (tasks not directly related to training neural networks, e.g., file I/O) as the volume of the training set decreases.

\begin{figure}[!htbp]
  \centering
  \includegraphics[trim={4.5cm 3.5cm 4.5cm 3.5cm},clip, width=.89\textwidth]{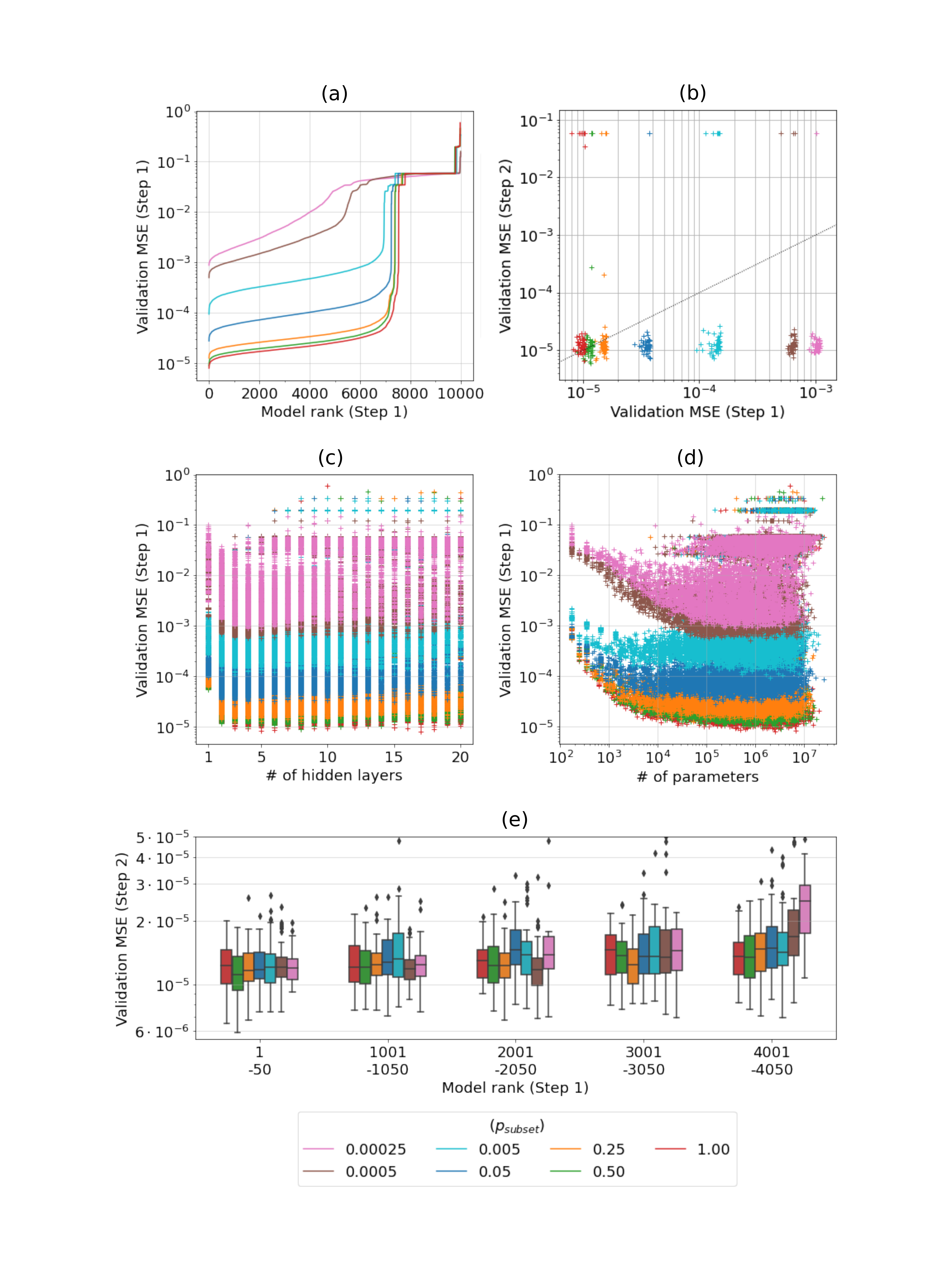}
  \caption{(Continued on the following page.)}
  \label{fig:fig2}
\end{figure}
\begin{figure}[t]
  \contcaption{(a) Minimum validation MSEs of all evaluation trials after sorted by their minimum validation MSE from Step 1. (b) Comparison between minimum validation MSE from Step 2 and minimum validation MSE from Step 1 for the top Step 1 candidate models (50 per HPO project).  (c and d) Relationship between the minimum validation MSE and the number of hidden layers and learnable parameters, respectively. (e) Step 1 models with different rank groups, e.g., rank 1–50, 1001–1050, 2001–2050, 3001–3050, and 4001–4050, are retrained with a full training dataset, and then their minimum validation MSE are displayed as box diagrams. Each rank group contains 50 models. The box shows the inter-quartile range (IQR) with a line at the median, and the whiskers 1.5-times the inter-quartile range (IQR). Flier points are outliers, i.e., those beyond the whiskers. Figures S4, S5, and S6 show similar figures as c, d, and e but in which each HPO project is plotted in a separate subpanel.}
\end{figure}

\section{Results}
An HPO project with a larger training dataset consistently yields trial models with lower validation MSEs after Step 1, except for ones that have markedly poor performance, e.g., validation MSE > 1×10$^{-2}$, (Figures 2a and S1). While it can be tempting to assume that data volume is all that matters, our working hypothesis is that just the top few architectures revealed in HPO projects with smaller subsets could perform just as well as the best architectures sampled at much greater expense in P1.00, once exposed to the full data in a second stage. One can imagine alternative views, such as an HPO with a smaller dataset being self-limiting in the sense that using a smaller dataset may result in different optima than an HPO with a larger dataset. However, we view this as unlikely since we use a random search algorithm in which the selection of trial hyperparameters is independent of the evaluations of preceding trials.

If our hypothesis is valid, we expect that top architectures harvested from HPO projects with smaller training datasets (P0.00025 to P0.50) shall have comparable performance to those from the HPO project that used the full training dataset (P1.00).  We verify this hypothesis in Step 2: the top 50 trial models from each HPO project are retrained with the full training dataset and then their performances are re-examined by comparing the validation MSE from Step 1 (using a subset) and that from Step 2 (using a full dataset). 

The results affirm our hypothesis: HPO with a small subset of a training dataset is found to be capable of discerning competent hyperparameter configurations (Figures 2b and S2). Despite the notable performance gaps across the four HPO projects in Step 1 (spreads across the x-axis in Figure 2b), the performances of top trial models of the subset HPO projects (P0.00025 to P0.50) after retraining generally converge to those from the P1.00 project within the same order of magnitude (convergence on the y-axis in Figure 2b). Figures 2b and S2 appear to exhibit a weak but noticeable sensitivity, e.g., increasing model errors as the subset size decreases; however, this is found to be an artifact due to the randomness introduced during training (further discussed in Section 6).  Admittedly, one cannot escape a few models that suffer degraded performance after retraining (i.e., all models with MSE larger than 1×10$^{-2}$ except in P0005 and P00025); model overfitting or stochastic aspects of training such as during weight initialization or shuffling samples before each epoch can cause such issues. This appears more prevalent in models with higher complexity, such as those with more hidden layers or more learnable parameters (Figure S3). But the main point is the reassuringly similar skill of all fits after retraining (Step 2) in Figure 2b, which affirms the two-step HPO method. In particular, it is remarkable that HPO with only a 0.025\% subset or $\sim$5K samples (P0.00025; cf. $\sim$20M samples in P1.00) is capable of identifying top-performing architectures.

Furthermore, the unusually \textit{extensive} HPO project performed here provides an opportunity to examine the effect of hyperparameters on the model performance. Figures 2c and 2d reveal a minimum model complexity required for faithfully emulating the given aerosol activation dataset: at least two or three hidden layers and 10$^4$–10$^5$ learnable parameters are required for optimal performance. Interestingly, the minimum learnable parameter for maximum performance seems to depend on the subset size, e.g., smaller for larger $p_{subset}$ ($\sim$10$^5$ for P0.00025 and $\sim$10$^4$ for P1.00). While it is not the main purpose of HPO, analyzing the relationship between hyperparameters and model performance provides useful information to understand model architectures and develop further refined models in the next iteration, though we admit it is undoubtedly problem-specific. 

\begin{figure}[!htbp]
  \centering
  \includegraphics[trim={17.5cm 9.1cm 17.5cm 8.9cm},clip, width=.95\textwidth]{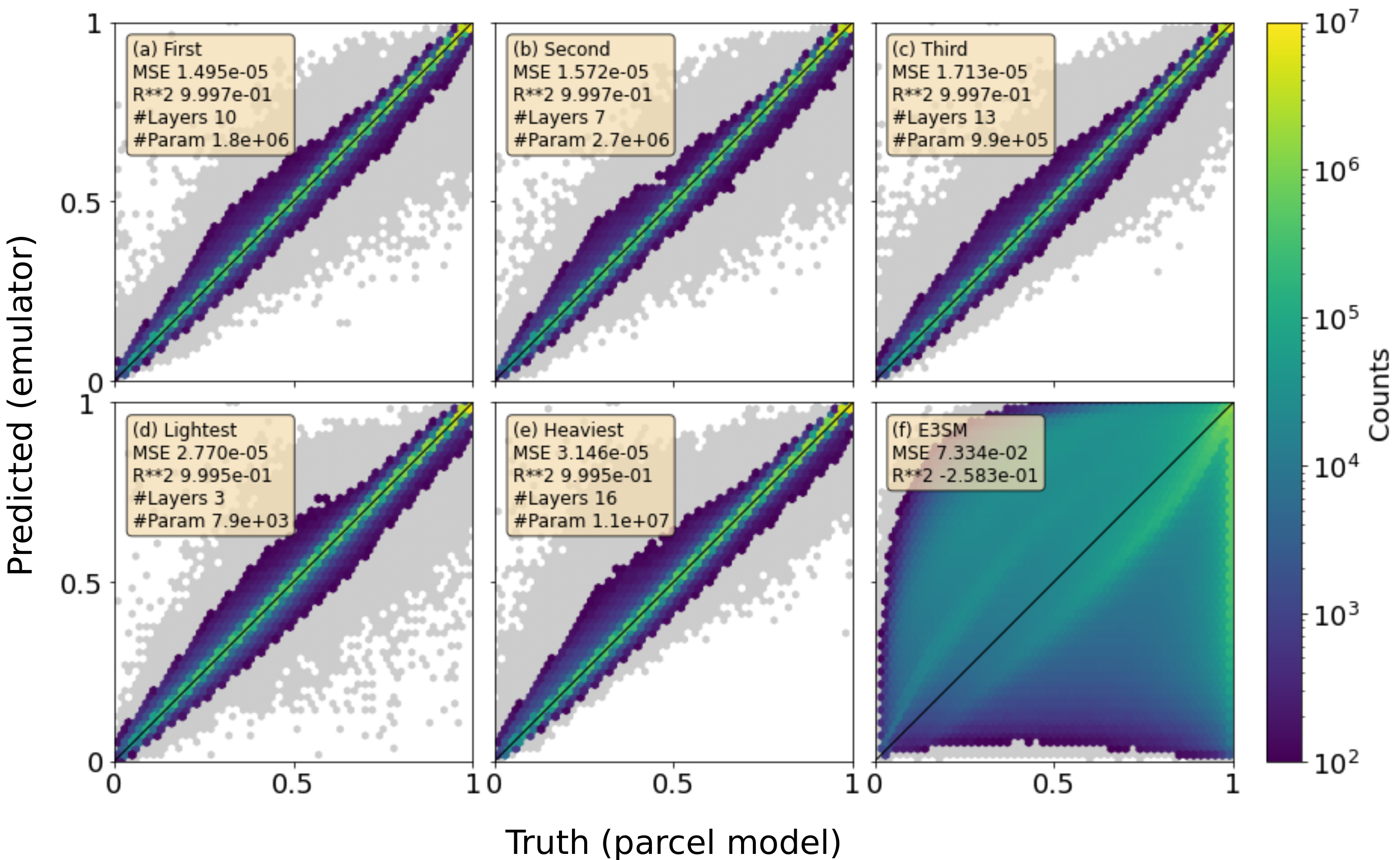}
  \caption{Hexagonally binned histogram with aerosol activation fractions (first mode only) computed by the cloud parcel model (x-axis) and estimated by different neural network emulators (y-axis). The subplots correspond to (a) the best emulator, (b) the second-best emulator, (c) the third-best emulator, (d) the lightest emulator, (e) the heaviest emulator, and (f) E3SM's current parameterization (Abdul‐Razzak and Ghan 2000). The total number of samples in this held-out test dataset is 7.9×10$^7$, and bins less than 100 samples (0.0001\% of the total sample) are shown in grey. Mean squared error (MSE) and coefficient of determination (R**2), the number of hidden layers (\#Layers), and the total number of learnable parameters (\#Param) are displayed in the top left corner of each subplot.}
  \label{fig:fig3}
\end{figure}

The accuracy of predictions by selected ML models after Step 2 is illustrated in Figure 3a–e by mapping the truth values (aerosol activation fractions calculated using the cloud parcel model) and the predicted values (predicted by a neural network emulator), using the holdout test dataset. The selected models from the entire Step 2 pool (total of 350 models) include the top three performers and the lightest and the heaviest models in terms of the number of learnable parameters. Note that the erroneous models whose MSE is larger than 1×10$^{-4}$ are excluded from the selection process. All five neural network models display stark improvement from the current aerosol activation parameterization used in E3SM (Figure 3f). However, the difference among them is barely noticeable despite large differences in model complexity—e.g., the number of learnable parameters of the five models spans three orders of magnitudes from 7.9×10$^3$ to 1.1×10$^7$. Remarkably, the lightest model (Figure 3d) performs almost as well as the best model (Figure 3a), which has about 1,400 times more learnable parameters. The \textit{extensive} HPO was key to finding this lucky architecture.

The diversity of resulting top-performing models is another key merit of an extensive HPO. As shown in Table S1, the model architectures of top-performing models vary significantly, providing users with a range of options depending on their specific application needs. For example, in the case of our aerosol activation emulators, the lightest model, though not the best performer, is a more strategic choice for inference within a climate model due to its computational efficiency given that it maintains performance (Figure S7).

\section{Setting parameters for two-step HPO}
Just like in any HPO method, our two-step HPO approach requires \textit{several} preset parameters:  $p_{subset}$ (the portion of training set used for Step 1 evaluation) and $p_{retrain}$ (the portion of Step 1 trials used for Step 2 re-evaluation) in addition to $n_{trials}$ (the number of trials; a parameter universally applicable for any HPO method). Here, we attempt to provide insights into choosing these parameters based on our case study. 

Figure 2a shows that model performance asymptotically converges after Step 1. For instance, for P0.005–P1.00, models with ranks below 6000 (“tier-1”) have validation MSE within the same order of magnitude. Given this outcome, the subsequent query that arises is how much model performance differs across the tier-1 models after retraining with the full training dataset. If performance is consistent for the tier-1 models post-retraining, a substantial value of $n_{trials}$ (e.g., 10,000) is not required to identify top candidates for Step 2 (e.g., 50 models) in our case study. This inquiry arises from our discovery that, among the top 50 models in each category, Step 1 ranks do not determine Step 2 ranks (Figure S8)—implying that the performance of models with equally competitive hyperparameter configurations is primarily governed by stochasticity during training (e.g., random weight initialization and mini-batch effects).

To address this question, we cluster Step1 models with different rank groups (e.g., rank 1–50, 1001–1050, 2001–2050, 3001–3050, and 4001–4050) and proceed to retrain them using the full training dataset (Step 2). We then make a box chart for the Step 2 validation MSEs per Step 1 rank group (see Figures 2e and S6). In the box chart, the Step 2 MSEs exhibit minimal variation among models belonging to different rank groups, except for P0.0005 and P0.00025 where Step 1 model performances do not show clear convergence (Figure 2a). This result indicates that we could have employed a significantly smaller $n_{trials}$ in Step 1 to identify the top 50 candidates for Step 2. For example, since we needed 50 top candidates in the asymptotically converged region, employing $n_{trials}$=100 for P0.005–P1.00 and $n_{trials}$=200 for P0.0005–P0.00025 would have sufficed, based on a random sampling of existing trials (Figure S9). However, to fully exploit the computational efficiency of this two-step HPO method, using a larger number of $n_{trials}$ is still beneficial, especially with smaller $p_{subset}$ values. In our case study, approximately 1,000 $n_{trials}$ is necessary to identify the relationship between hyperparameters and model performance (Figures S10 and S11).

This retrospective analysis points out the potential utility of a model performance versus rank diagram (akin to Figure 2a) to ascertain the adequacy of the three parameters within the two-step HP method.  An effective parameter configuration should yield asymptotic convergence in the performance–rank diagram, and $n_{trials}$ should be large enough that the number of top candidates for Step 2 ($p_{retrain} \cdot n_{trials}$) falls within the region of asymptotic convergence. However, further investigation and additional case studies are still required to attain more thorough guidance on the choice of two-step HPO parameters.

\section{Summary}
We have shared our lesson learned that a two-step approach is an attractive way to efficiently optimize hyperparameters in extensive architecture searches. In the first step, we perform HPO with a small subset of a training dataset to identify a set of hyperparameter combinations, i.e., viable architectures. In the second step, we retrain the top-performing configurations with the entire training dataset for the final selection and the recalibration of model weights. Our case study of optimizing neural network emulators for aerosol activation showed that the two-step HPO with only 0.025\% of the training dataset was effective in finding optimal hyperparameter configurations. Our case study, which is to our knowledge the first in-depth application of the core-set approach popularized in computer science literature (e.g., Feldman (2020)) to an operational climate modeling parameterization research problem, confirms the usefulness of the core-set approach for problems with a wide range of data volumes (e.g., 5K to 20M samples as shown in our case study). Additionally, we demonstrated other benefits of extensive HPO (i.e., many trials with large hyperparameter search domains). For example, an extensive HPO provides an opportunity to learn the relationship between hyperparameters and model performance. Moreover, it offers the option to choose the \textit{right} hyperparameter configuration amongst comparable options, in the balance between fit precision and inference efficiency that is relevant to hybrid ML-physics climate modeling.

We hope that this two-step HPO method, and appendices illustrating its practical use on GPU clusters, will prove a practical solution for ML practitioners who crave the benefits of extensive HPO, by reducing the time and compute involved. Additionally, this method can be easily tailored to specific needs. For example, the selection of top candidates for the second step can be modified (e.g., choosing only ones with two or three hidden layers if a compact architecture is desired for inference efficiency), or the second step could be another HPO process but with a more focused search space. 

We note that our two-step HPO method has only been demonstrated using a random search algorithm and its applicability with other search algorithms remains to be explored. Despite this limitation, the potential cost savings offered by the two-step approach with random search may outweigh the benefits of using an adaptive algorithm. Random searches are attractively parallel and can scale across GPU nodes. Determining the optimal values for the parameters required for the two-step HPO method ($n_{trials}$, $p_{subset}$, and $p_{retrain}$) is also an open question that warrants further research. In our case study, $n_{trials}$=10000, $p_{subset}$= 0.00025, and $p_{retrain}$=0.005 were effective, but the optimal values may vary depending on the complexity of a training dataset and the expansiveness of a hyperparameter search space.

We hope others may share lessons learned from applying similar techniques in other settings. Machine learning parameterization is an empirical art, and much is yet to be learned from dense empirical sampling. Meanwhile, independent testing of this method has enabled competitive results beyond the aerosol nucleation use case discussed here, in a new benchmark project focused on ML parameterization for convection and radiation parameterization (Yu et al. 2023).

\section*{Acknowledgments}
We thank Elizabeth A. Barnes and two anonymous reviewers for providing valuable feedback that improved our manuscript significantly as well as our UCI colleagues Savannah Ferretti, Jerry Lin, and Yan Xia for helpful comments. This study was supported by the Enabling Aerosol-cloud interactions at Global convection-permitting scalES (EAGLES) project (project no. 74358), funded by the U.S. Department of Energy (DOE), Office of Science, Office of Biological and Environmental Research, Earth System Model Development (ESMD) program area. This research used resources of the National Energy Research Scientific Computing Center (NERSC), a U.S. DOE Office of Science User Facility located at Lawrence Berkeley National Laboratory, operated under Contract No. DE-AC02-05CH11231 using NERSC awards ALCC-ERCAP0016315, BER-ERCAP0015329, BER-ERCAP0018473, and BER-ERCAP0020990. This research also used resources of the Argonne Leadership Computing Facility, a DOE Office of Science User Facility supported under Contract DE-AC02-06CH11357, using an award of computer time provided by the Advanced Scientific Computing Research (ASCR) Leadership Computing Challenge (ALCC) program. The Pacific Northwest National Laboratory is operated for the U.S. DOE by Battelle Memorial Institute under contract DE-AC05-76RL01830. Additionally, this work used XSEDE’s PSC Bridges-2 system (National Science Foundation grant ACI-1548562 and ACI-1928147). We thank David Walling for his assistance with HPC support through the XSEDE Extended Collaborative Support Service program.

\section*{Data Availability Statement}
The aerosol activation dataset used in this study is openly available on the Zenodo data repository (\url{https://doi.org/10.5281/zenodo.7627577}). The two-step hyperparameter optimization codes used in this study are openly available on the GitHub repository (\url{https://github.com/sungdukyu/Two-step-HPO}) and described in the Supplemental Material.

\section*{References} 
\begin{hangparas}{.25in}{1}
Abdul‐Razzak, H., and S. J. Ghan, 2000: A parameterization of aerosol activation: 2. Multiple aerosol types. J Geophys Res Atmospheres, 105, 6837–6844, https://doi.org/10.1029/1999jd901161.

Alfonso, L., and J. M. Zamora, 2021: A two-moment machine learning parameterization of the autoconversion process. Atmos Res, 249, 105269, https://doi.org/10.1016/j.atmosres.2020.105269.

Bergstra, J., and Y. Bengio, 2012: Random Search for Hyper-Parameter Optimization. Journal of Machine Learning Research, 13, 281–305.

Chiu, J. C., C. K. Yang, P. J. van Leeuwen, G. Feingold, R. Wood, Y. Blanchard, F. Mei, and J. Wang, 2021: Observational Constraints on Warm Cloud Microphysical Processes Using Machine Learning and Optimization Techniques. Geophys Res Lett, 48, e2020GL091236, https://doi.org/10.1029/2020gl091236.

Feldman, D., 2020: Core-sets: an Updated Survey. arXiv, 10, https://doi.org/10.48550/arxiv.2011.09384.
Gettelman, A., D. J. Gagne, C. ‐C. Chen, M. W. Christensen, Z. J. Lebo, H. Morrison, and G. Gantos, 2021: Machine Learning the Warm Rain Process. J Adv Model Earth Sy, 13, https://doi.org/10.1029/2020ms002268.

Ghan, S. J., and Coauthors, 2011: Droplet nucleation: Physically‐based parameterizations and comparative evaluation. J Adv Model Earth Sy, 3, https://doi.org/10.1029/2011ms000074.

Golaz, J., and Coauthors, 2019: The DOE E3SM Coupled Model Version 1: Overview and Evaluation at Standard Resolution. J Adv Model Earth Sy, 11, 2089–2129, https://doi.org/10.1029/2018ms001603.

Killamsetty, K., D. Sivasubramanian, G. Ramakrishnan, A. De, and R. Iyer, 2021a: GRAD-MATCH: Gradient Matching based Data Subset Selection for Efficient Deep Model Training. arXiv, https://doi.org/10.48550/arxiv.2103.00123.

——, ——, ——, and R. Iyer, 2021b: GLISTER: Generalization based Data Subset Selection for Efficient and Robust Learning. Proc. AAAI Conf. Artif. Intell., 35, 8110–8118, https://doi.org/10.1609/aaai.v35i9.16988.

——, G. S. Abhishek, Aakriti, A. V. Evfimievski, L. Popa, G. Ramakrishnan, and R. Iyer, 2022: AUTOMATA: Gradient Based Data Subset Selection for Compute-Efficient Hyper-parameter Tuning. arXiv, https://doi.org/10.48550/arxiv.2203.08212.
Kingma, D. P., and J. Ba, 2014: Adam: A Method for Stochastic Optimization. Arxiv, https://doi.org/10.48550/arxiv.1412.6980.

Li, L., K. Jamieson, G. DeSalvo, A. Rostamizadeh, and A. Talwalkar, 2016: Hyperband: A Novel Bandit-Based Approach to Hyperparameter Optimization. Arxiv, https://arxiv.org/abs/1603.06560.

Liu, X., P.-L. Ma, H. Wang, S. Tilmes, B. Singh, R. C. Easter, S. J. Ghan, and P. J. Rasch, 2016: Description and evaluation of a new four-mode version of the Modal Aerosol Module (MAM4) within version 5.3 of the Community Atmosphere Model. Geosci. Model Dev., 9, 505–522, https://doi.org/10.5194/gmd-9-505-2016.

Mirzasoleiman, B., J. Bilmes, and J. Leskovec, 2020: Coresets for Data-efficient Training of Machine Learning Models. arXiv, https://doi.org/10.48550/arxiv.1906.01827.

Rasch, P. J., and Coauthors, 2019: An Overview of the Atmospheric Component of the Energy Exascale Earth System Model. J. Adv. Model. Earth Syst., 11, 2377–2411, https://doi.org/10.1029/2019ms001629.

Silva, S. J., P.-L. Ma, J. C. Hardin, and D. Rothenberg, 2021: Physically regularized machine learning emulators of aerosol activation. Geosci Model Dev, 14, 3067–3077, https://doi.org/10.5194/gmd-14-3067-2021.

Snoek, J., H. Larochelle, and R. P. Adams, 2012: Practical Bayesian Optimization of Machine Learning Algorithms. arXiv, https://doi.org/10.48550/arxiv.1206.2944.

Towns, J., and Coauthors, 2014: XSEDE: Accelerating Scientific Discovery. Comput Sci Eng, 16, 62–74, https://doi.org/10.1109/mcse.2014.80.

Visalpara, S., K. Killamsetty, and R. Iyer, 2021: A Data Subset Selection Framework for Efficient Hyper-Parameter Tuning and Automatic Machine Learning. SubSetML Workshop 2021 at the International Conference on Machine Learning.

Wang, H., and Coauthors, 2020: Aerosols in the E3SM Version 1: New Developments and Their Impacts on Radiative Forcing. J. Adv. Model. Earth Syst., 12, https://doi.org/10.1029/2019ms001851.

Wendt, A., M. Wuschnig, and M. Lechner, 2020: Speeding up Common Hyperparameter Optimization Methods by a Two-Phase-Search. Iecon 2020 46th Annu Conf Ieee Industrial Electron Soc, 00, 517–522, https://doi.org/10.1109/iecon43393.2020.9254801.

Yu, S., and Coauthors, 2023: ClimSim: An open large-scale dataset for training high-resolution physics emulators in hybrid multi-scale climate simulators. arXiv, https://doi.org/10.48550/arxiv.2306.08754.
\end{hangparas}

\pagebreak

\begin{center}
  \textbf{\LARGE Supplemental Material}
\end{center}
\setcounter{figure}{0}
\setcounter{table}{0}
\renewcommand{\thetable}{S\arabic{table}}
\renewcommand{\thefigure}{S\arabic{figure}}

\begin{figure}[htp]
  \centering
  \includegraphics[trim={18.5cm .5cm 18.5cm .5cm},clip, width=.9\textwidth]{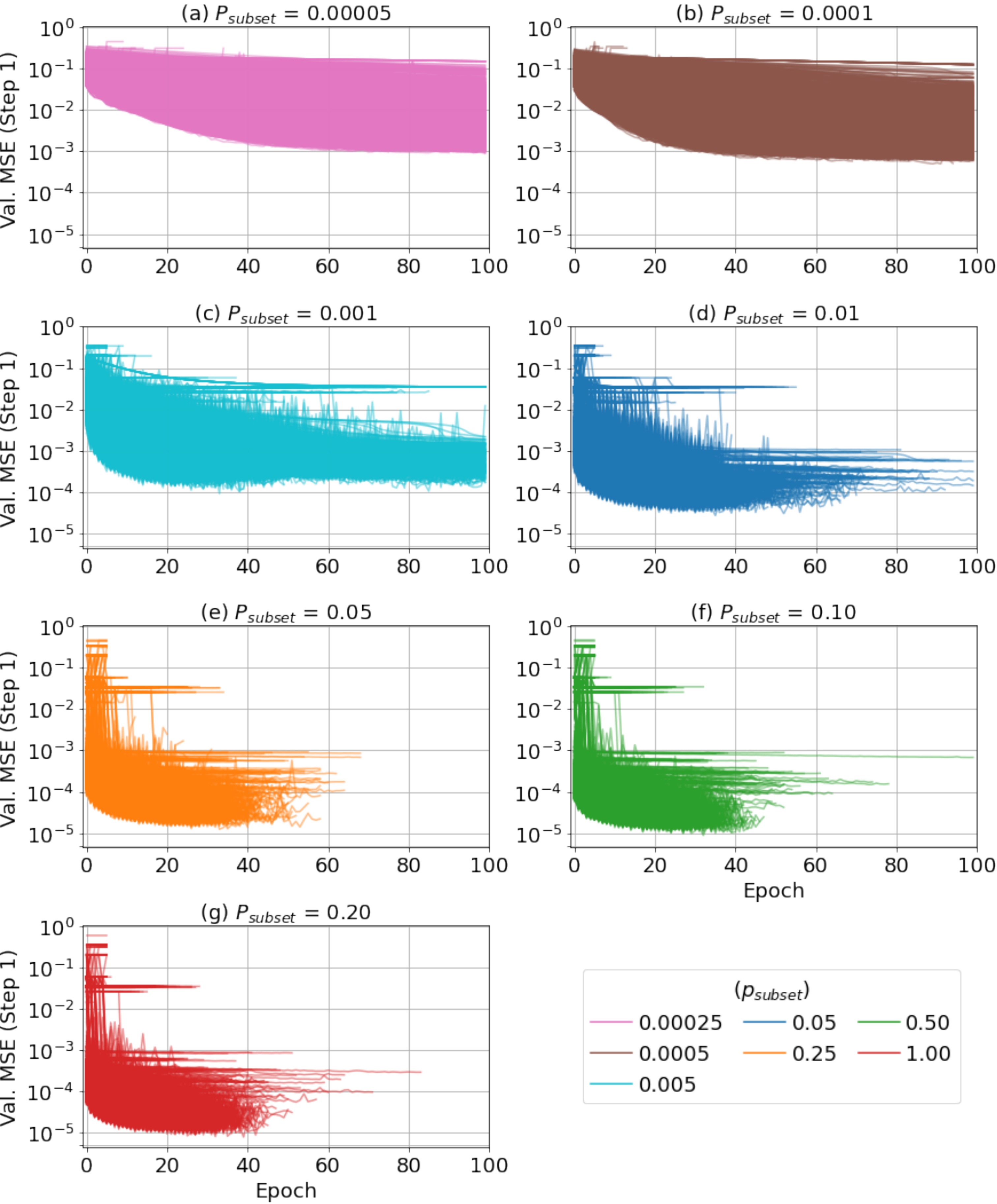}
  \caption{Validation MSE over training epochs during Step 1. Each line presents one individual tuning trial. Note that an early stopping rule is enforced with a patience of 5 epochs, while the maximum training epoch number is 100.}
\end{figure}

\begin{figure}[htp]
  \centering
  \includegraphics[trim={23cm 12cm 22cm 12cm},clip, width=.775\textwidth]{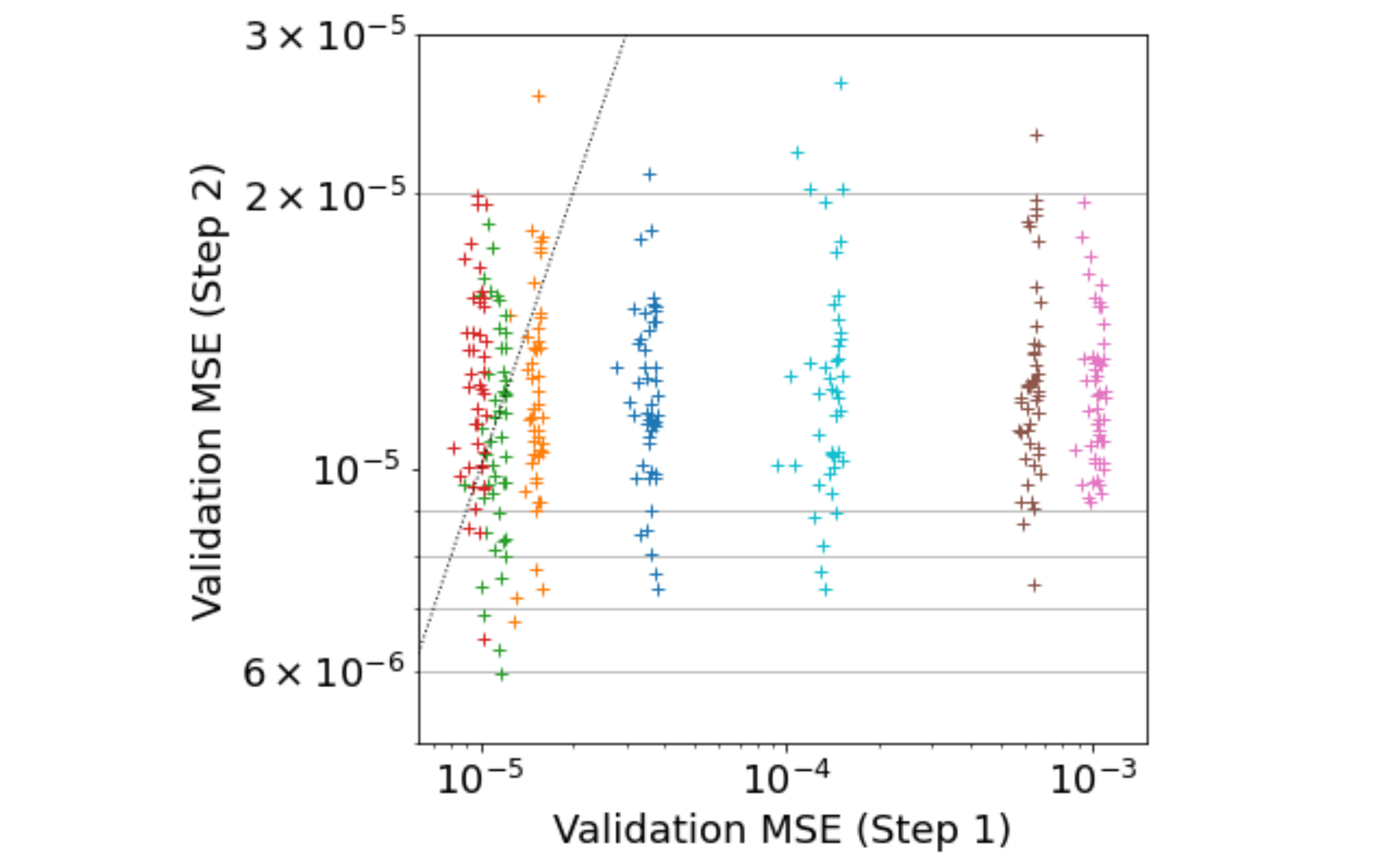}
  \caption{Same as Figure 2b, except the y-axis view limits.}
\end{figure}

\begin{figure}[htp]
  \centering
  \includegraphics[trim={18.5cm 14cm 18.5cm 13cm},clip, width=.95\textwidth]{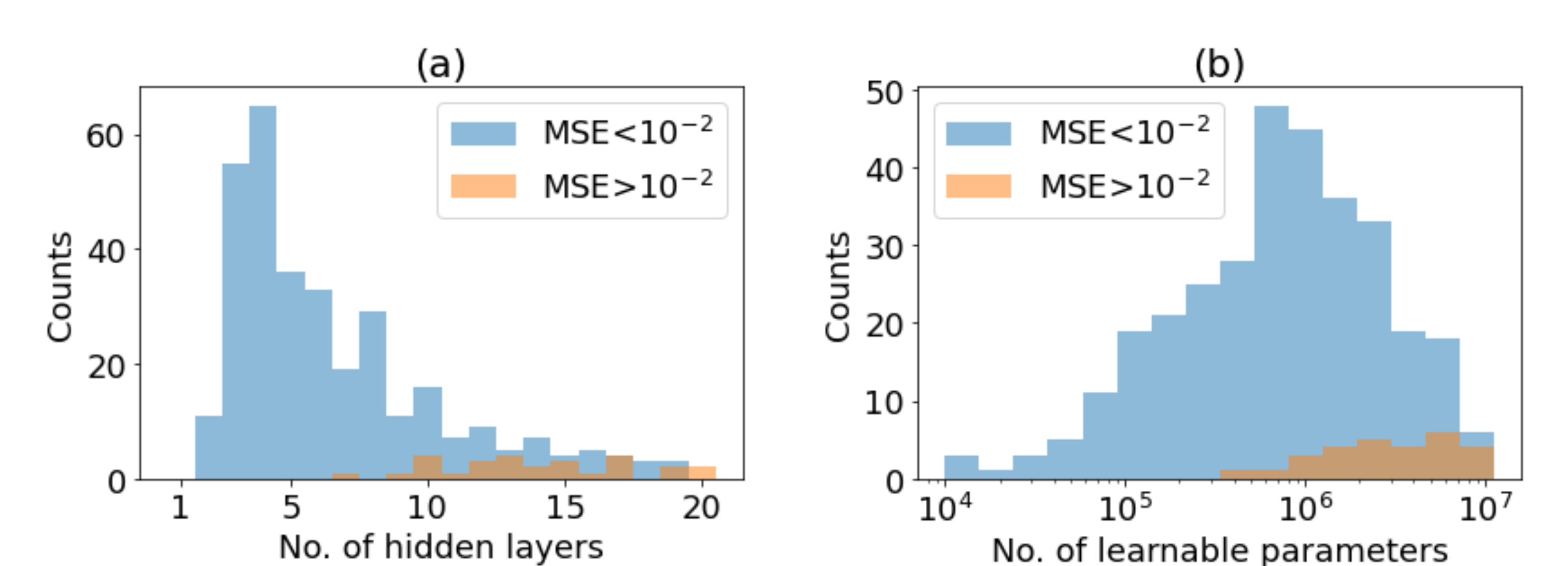}
  \caption{Histogram of (a) the number of hidden layers and (b) the number of learnable parameters of 350 trial models (top 50 models from seven HPO projects) after retraining in Step 2. Blue (orange) bars show models with their test MSE lower than (larger than) 1×10$^{-2}$.}
\end{figure}

\begin{figure}[htp]
  \centering
  \includegraphics[trim={18.5cm .5cm 18.5cm .5cm},clip, width=.9\textwidth]{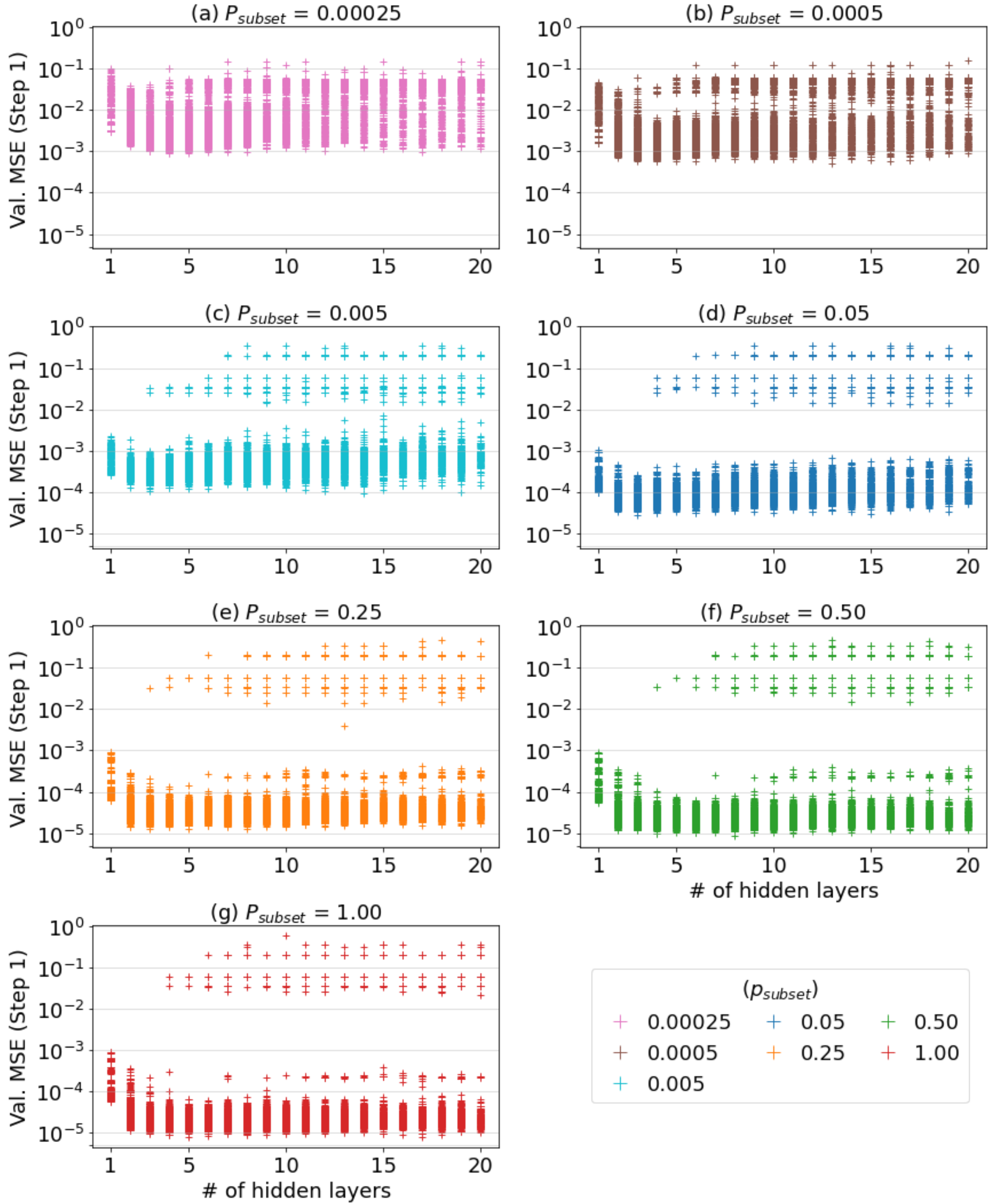}
  \caption{Same as Figure 2c, except that each HPO project is plotted in a separate subpanel.}
\end{figure}

\begin{figure}[htp]
  \centering
  \includegraphics[trim={18.5cm .5cm 18.5cm .5cm},clip, width=.9\textwidth]{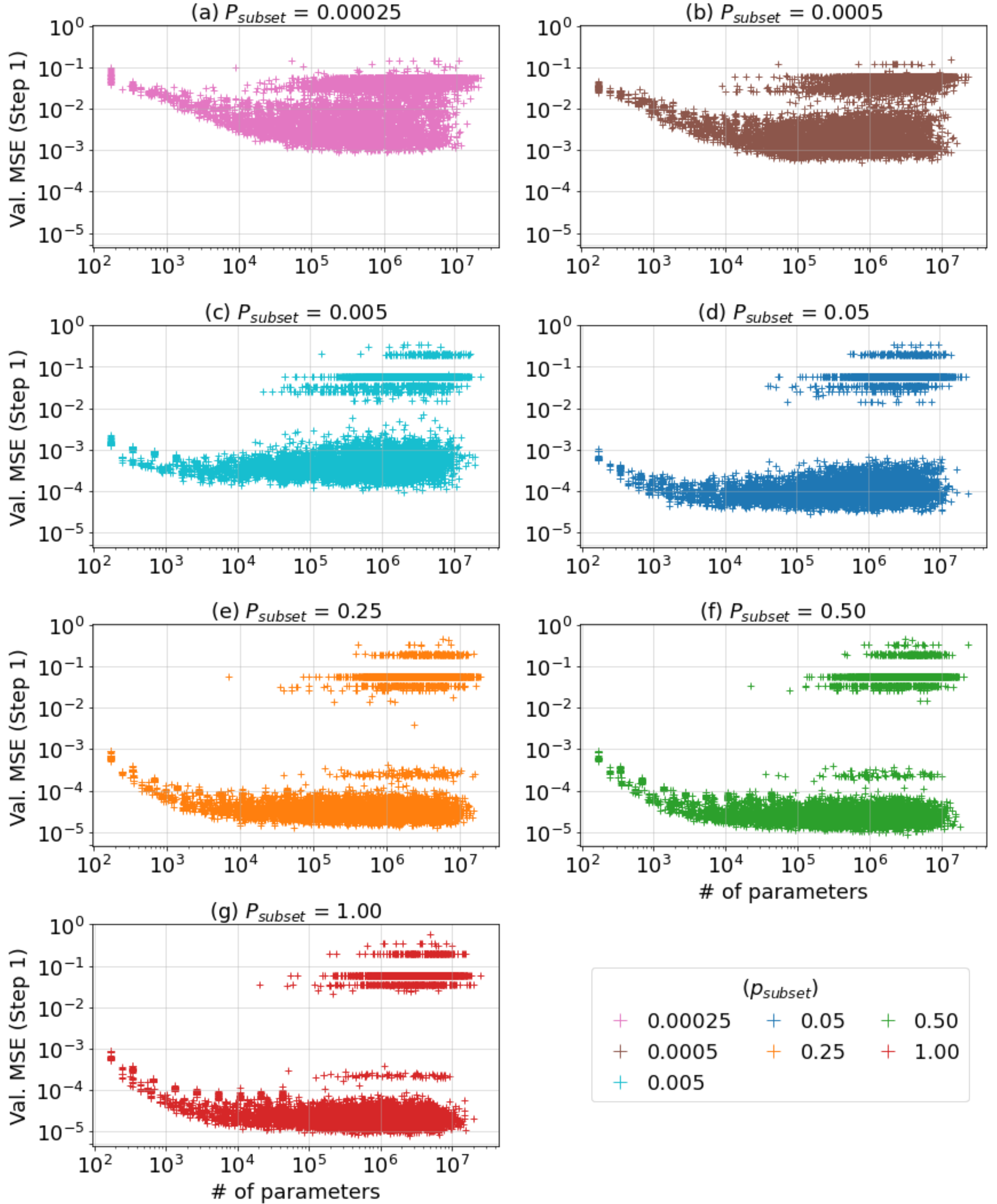}
  \caption{Same as Figure 2d, except that each HPO project is plotted in a separate subpanel.}
\end{figure}

\begin{figure}[htp]
  \centering
  \includegraphics[trim={17.5cm 9.2cm 17.5cm 9.2cm},clip, width=.9\textwidth]{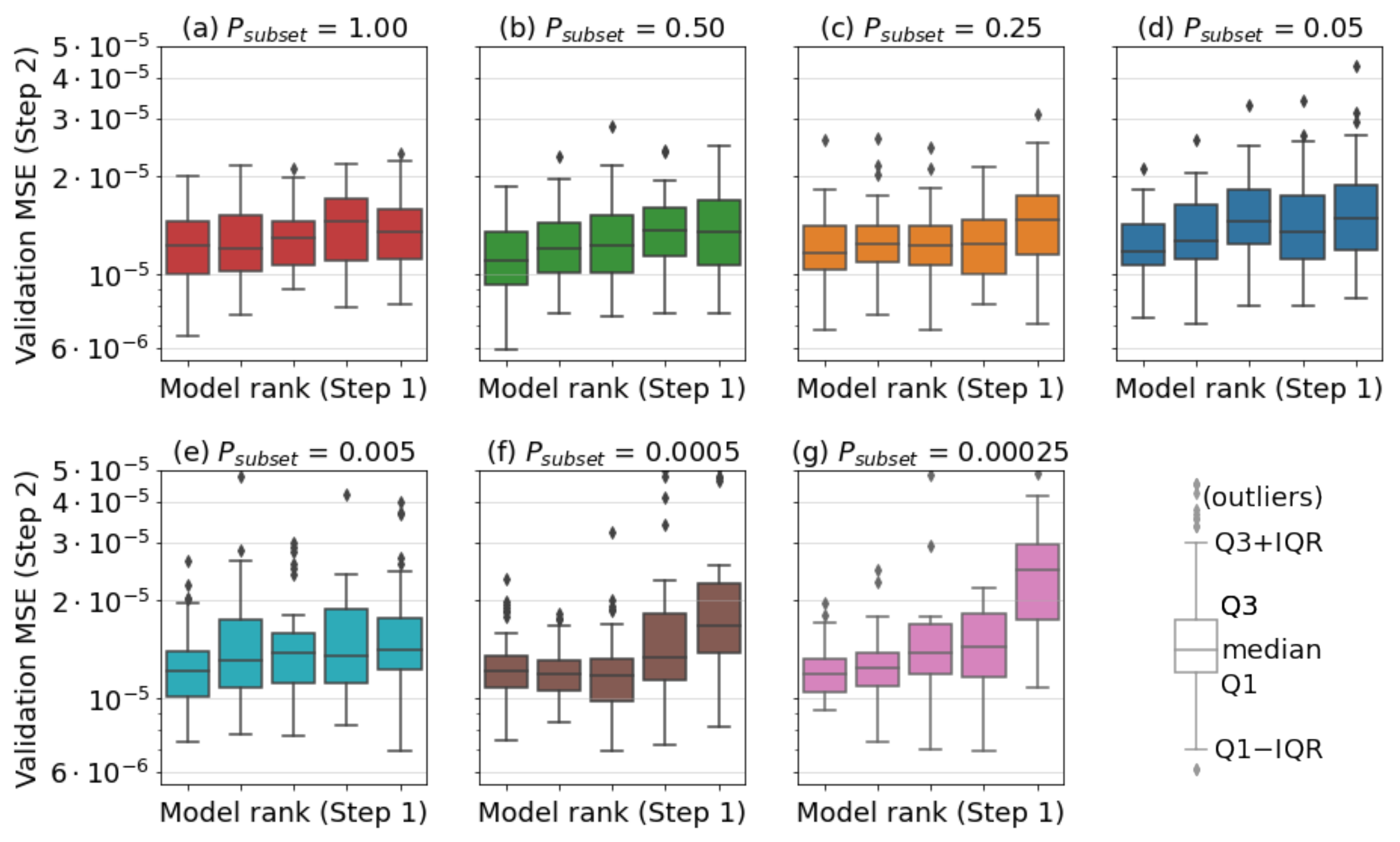}
  \caption{Same as Figure 2e, except that each HPO project is plotted in a separate subpanel.}
\end{figure}

\begin{figure}[htp]
  \centering
  \includegraphics[trim={23cm 12cm 22cm 12cm},clip, width=.75\textwidth]{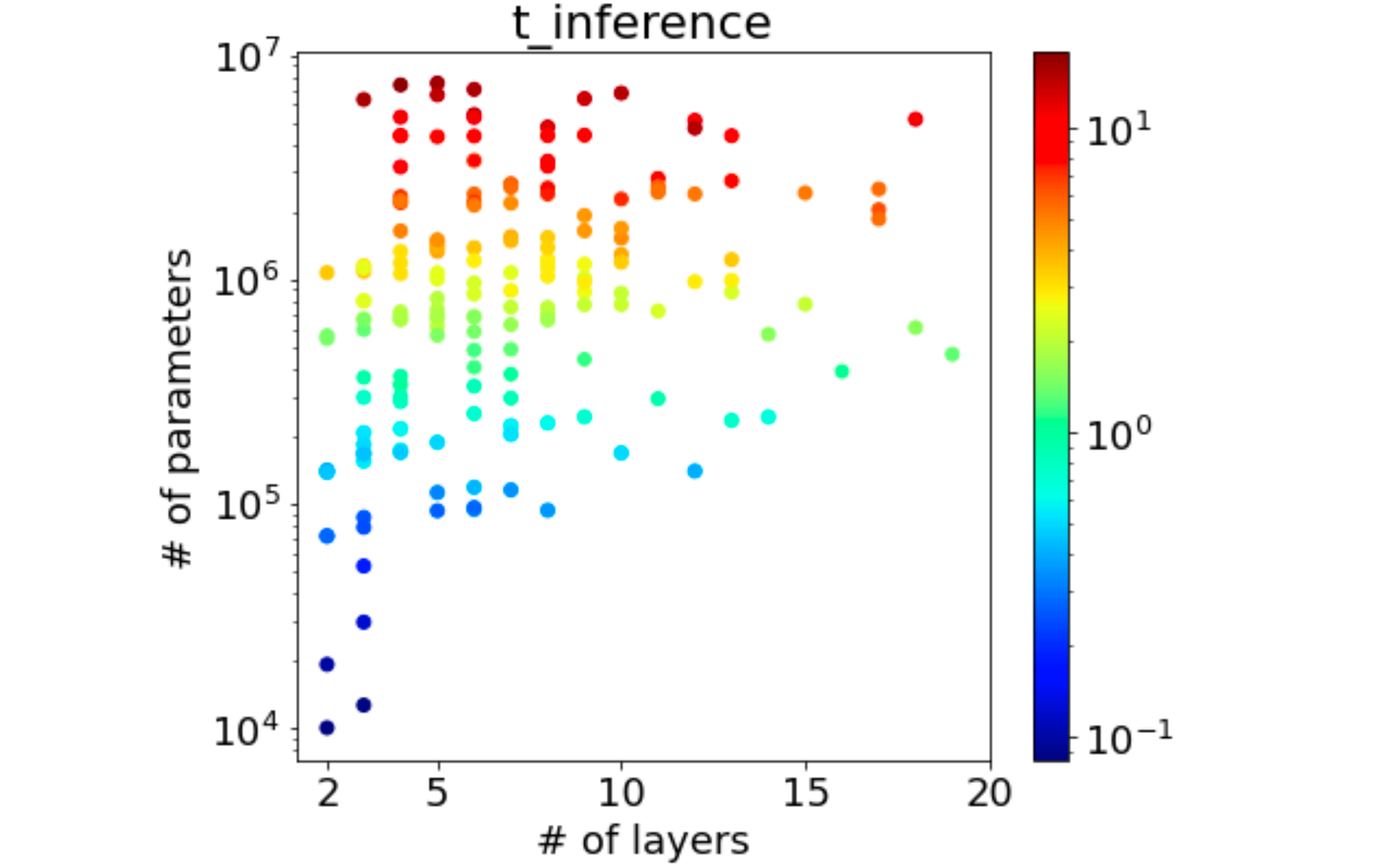}
  \caption{Inference time for 100,000 samples with a batch size of 1,024. Inference was repeated 10 times, and then, the inference times were averaged. One CPU core (AMD EPYC 7742) is used. Unit: seconds.}
\end{figure}

\begin{figure}[htp]
  \centering
  \includegraphics[trim={19cm 8.5cm 19cm 8.5cm},clip, width=.85\textwidth]{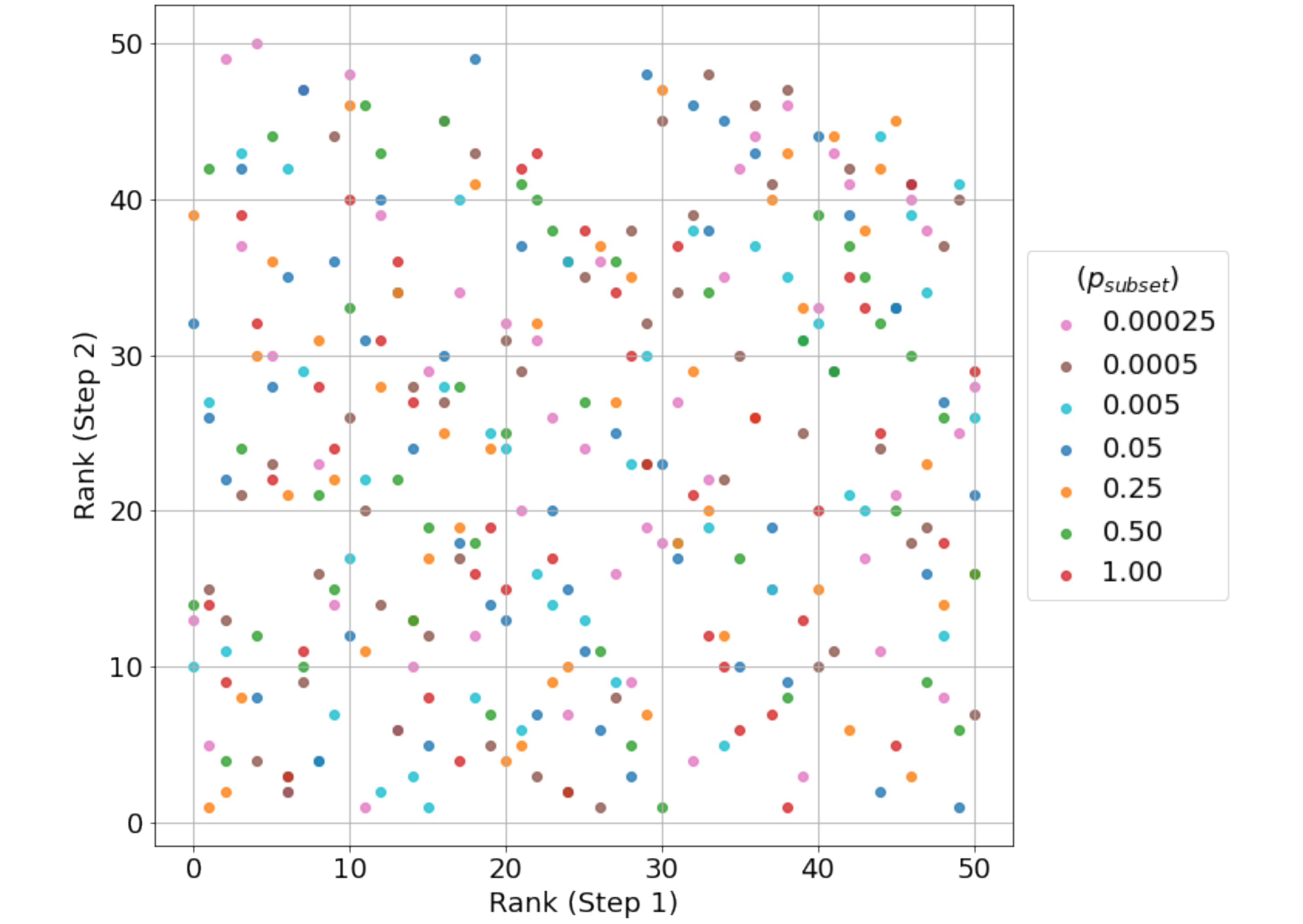}
  \caption{Scatter plot between Step-1 ranks (x-axis) and Step-2 ranks (y-axis) for top-50 models of each HPO project.}
\end{figure}

\begin{figure}[htp]
  \centering
  \includegraphics[trim={17.5cm 1.25cm 15.1cm 1.25cm},clip, width=.9\textwidth]{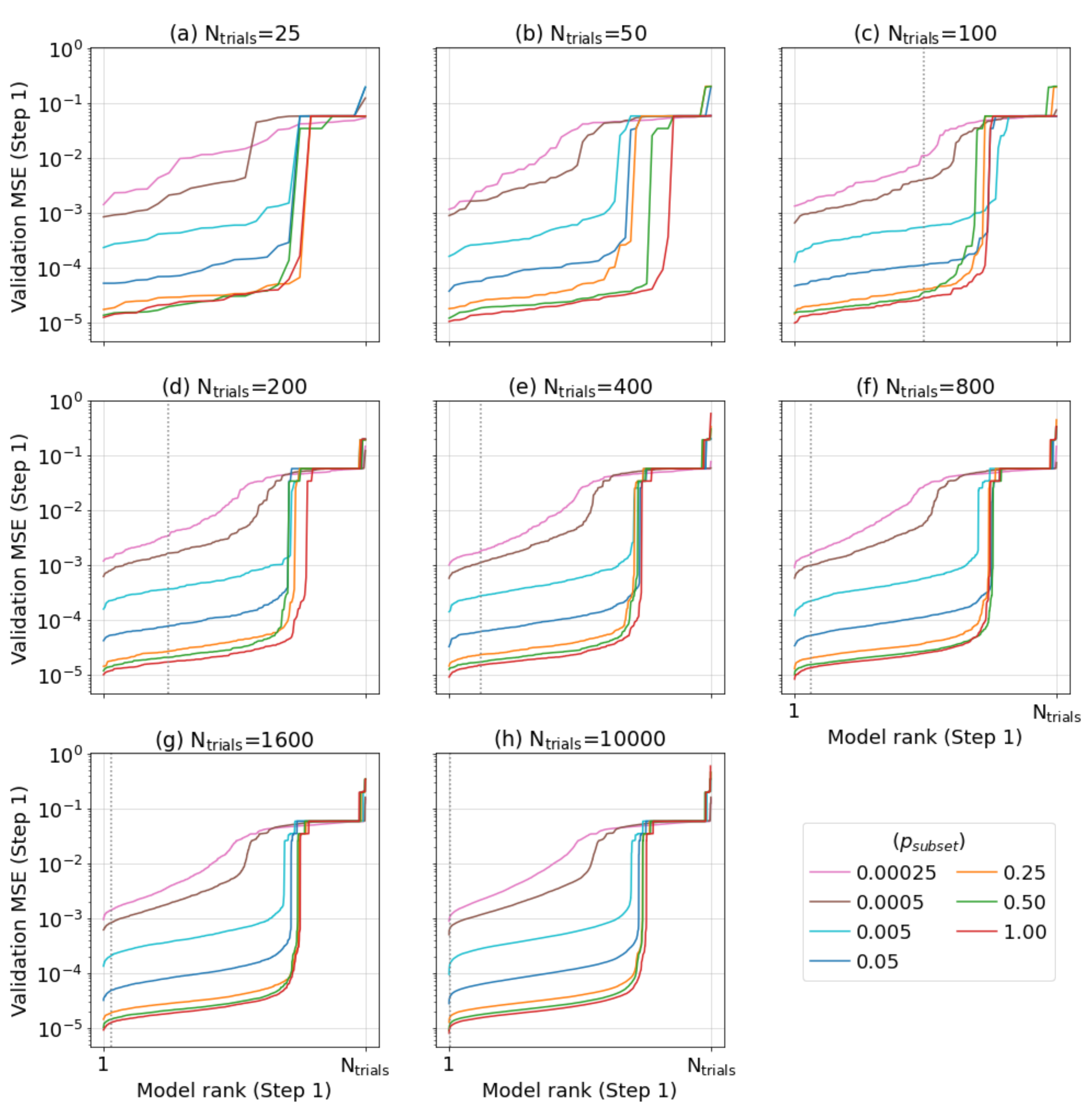}
  \caption{Same as Figure 2a, except that it is drawn after randomly drawn trials without replacement. The number of trials used in each panel is 25 (a), 50 (b), 100 (c), 200 (d), 400 (e), 800 (f), 1600 (g), and 10000 (h). Panel (h) is exactly identical to Figure 2a. Dashed lines denote the x value of 50.}
\end{figure}

\begin{figure}[htp]
  \centering
  \includegraphics[trim={17.5cm 1.25cm 15.1cm 1.25cm},clip, width=.9\textwidth]{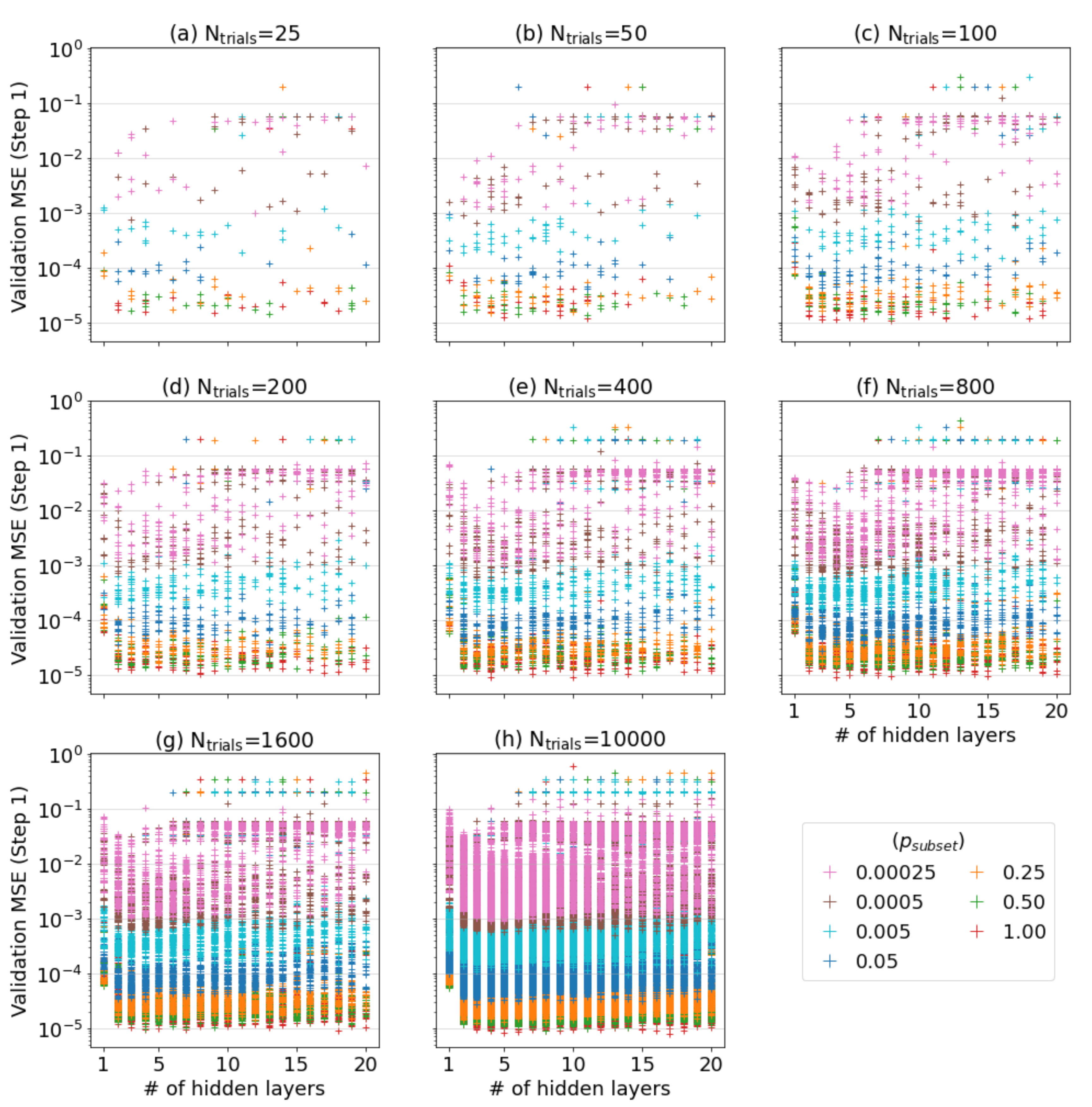}
  \caption{Same as Figure 2c, except that it is drawn after randomly drawn trials without replacement. The number of trials used in each panel is 25 (a), 50 (b), 100 (c), 200 (d), 400 (e), 800 (f), 1600 (g), and 10000 (h). Panel (h) is identical to Figure 2c.}
\end{figure}

\begin{figure}[htp]
  \centering
  \includegraphics[trim={17.5cm 1.25cm 15.1cm 1.25cm},clip, width=.9\textwidth]{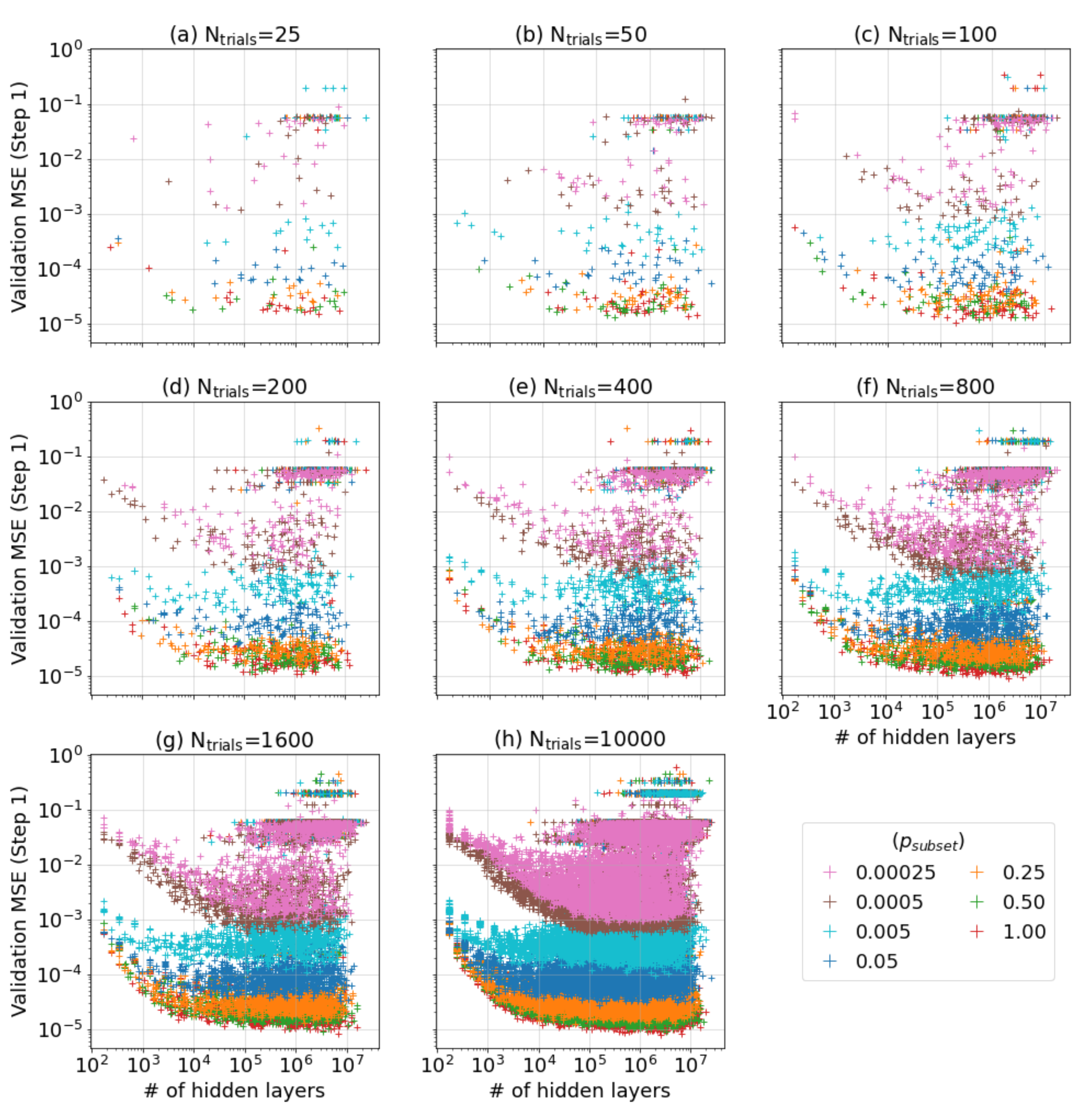}
  \caption{Same as Figure 2d, except that it is drawn after randomly drawn trials without replacement. The number of trials used in each panel is 25 (a), 50 (b), 100 (c), 200 (d), 400 (e), 800 (f), 1600 (g), and 10000 (h). Panel (h) is identical to Figure 2d.}
\end{figure}

\clearpage

\begin{table}[htp]
\centering
\includegraphics[trim={2.5cm 4cm 2.5cm 3cm},clip, width=\textwidth]{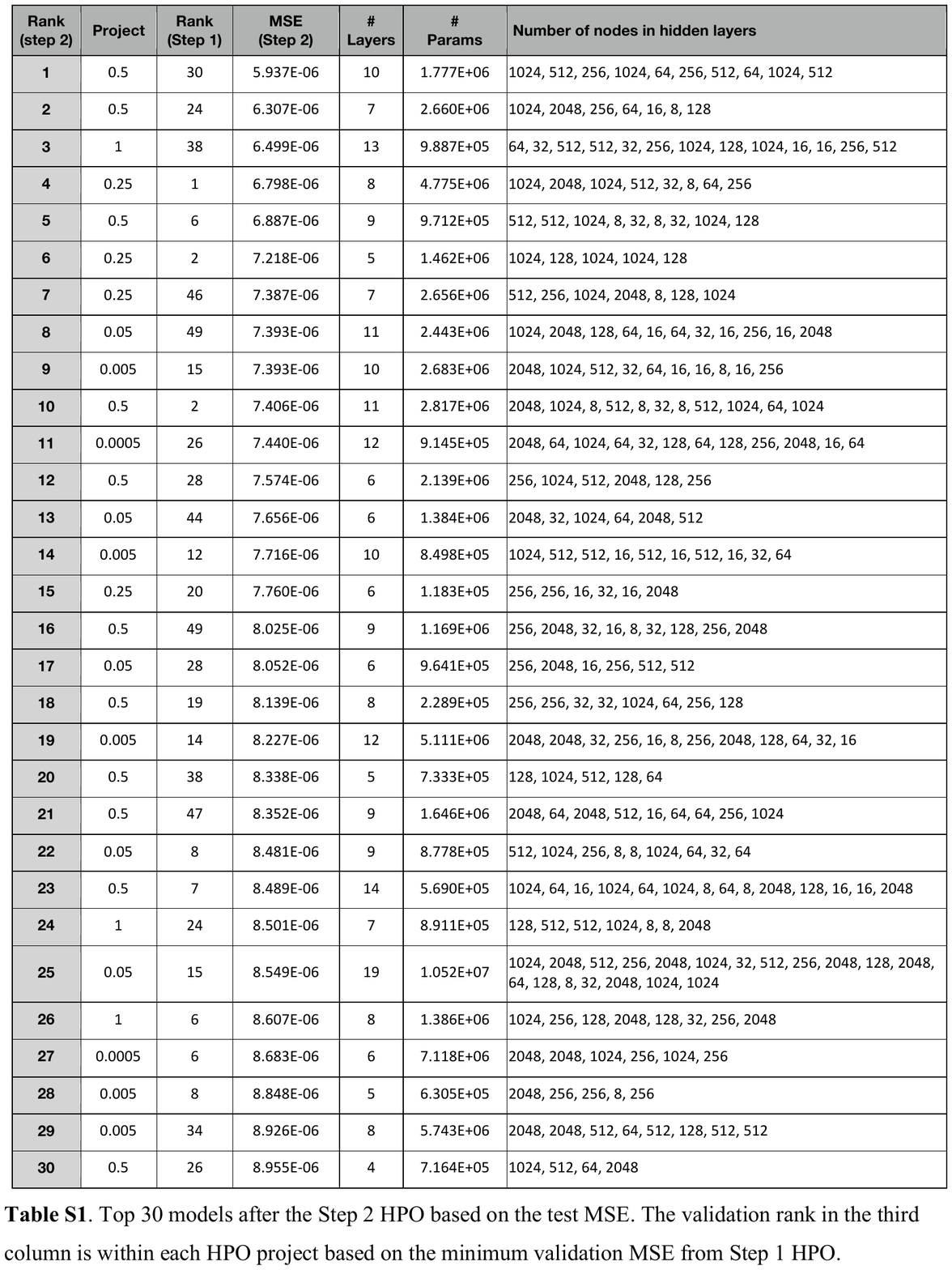}
\caption{Top 30 models after the Step 2 HPO based on the test MSE. The validation rank in the third column is within each HPO project based on the minimum validation MSE from Step 1 HPO.}
\end{table}

\clearpage

\textbf{Text S1}. Code example to set up a distributed mode in Keras Tuner in a SLURM-managed high-performance computing cluster (HPC).
\vspace{12pt}

Keras Tuner provides a high-level interface that enables distributed search mode with only four environmental variables. These include three Keras Tuner-specific variables (1-3) and one CUDA-related variable (4):
\begin{enumerate}
\item KERASTUNER\_TUNER\_ID: A unique ID assigned to manager and worker processes, with "chief" used for the manager process.
\item KERASTUNER\_ORACLE\_IP: IP address or hostname of the manager (chief) process.
\item KERASTUNER\_ORACLE\_PORT: Port number of the manager (chief) process.
\item CUDA\_VISIBLE\_DEVICES: Local GPU device ID to control the number and selection of GPUs assigned to a single process.
\end{enumerate}
These four environmental variables are assigned dynamically depending on the availability of computing resources in an HPC. Accordingly, it can be technically challenging to write scripts that automatically set up correct values for the above four environmental variables. To address this, we provide code examples based on the scripts used in our case study.
 
Our case study was conducted on PSC Bridges-2 (https://www.psc.edu/resources/bridges-2/), which has 8 GPUs per node in its GPU partition. We used two nodes (16 GPUs in total) and assigned one GPU to each worker process. To automatically set the above four environmental variables, we wrote three scripts:

-       Script 1. <sbatch-keras-tuner.sh>
: SLRUM job submission script that launches parallel jobs (in our case, ‘run-dynamic.sh’) using srun. The following options need to be set based on a user’s needs:

\begin{itemize}
\item   partition: name of a GPU partition
\item   nodes: number GPU nodes
\item   gpus: number of total GPUs processes, i.e., nodes x GPUs/node
\item   ntasks: number of tasks, i.e., gpus +1. Extra one process accounts for the process for a manager. 
\end{itemize} 
-       Script 2. <run-dynamic.sh>
: Intermediary script bridging the SLURM script (‘sbatch-kears-tuner.sh’) and the python script using Keras Tuner (‘keras-tuner-dynamic.py’). It is launched in parallel with each job step having its own SLURM-generated environmental variables, e.g., SLURM\_LOCALID and SLURMD\_NODENAME.
 
-       Script 3. <keras-tuner-dyanamic.py>
: The python script for hyperparameter tuning using Keras Tuner, with the environmental variables (1-4) set automatically based on SLURM variables.:
\begin{itemize}
\item   num\_gpus\_per\_node (Line 1): the number of GPUs per node.
\end{itemize}

\vspace{12pt}
\lstset{language=bash}
\begin{center}
  \textbf{Script 1.} sbatch-keras-tuner.sh
\end{center}
\begin{lstlisting}
#!/bin/sh
 
## The following options should be supplied by a user
#SBATCH --job-name=
#SBATCH --output=
#SBATCH --account=
#SBATCH --time=
 
## The following options should also be supplied by a user.
## However, we provide an example of using 2 nodes in PSC Bridges-2
 
#SBATCH --partition=GPU # Name of Bridges-2's gpu partition
#SBATCH --nodes=2       # Number of nodes
#SBATCH --gpus=16       # Total number of gpu processors:
                        # i.e., <Number of nodes> times <Number of GPUs per node>
#SBATCH --ntasks=17     # Total number of tasks:
                        # should be total number of GPUs plus one,
                        # i.e., <Number of nodes> times <Number of GPUs per node> + 1
 
srun --wait=0 bash run-dynamic.sh
\end{lstlisting}

\begin{center}
  \textbf{Script 2.} run-dynamic.sh
\end{center}
\begin{lstlisting}
#!/bin/sh
 
# The following three lines are optional for a log file.
# SLURM_LOCALID and SLURMD_NODENAME are environmental variables that automatically set by SLURM.
echo "--- run-dynamic.sh ---"
echo SLURM_LOCALID   $SLURM_LOCALID
echo SLURMD_NODENAME $SLURMD_NODENAME
 
# Run the keras-tuner script.
# Standard output and errors are printed to an inidividual log file.
python keras-tuner-dynamic.py > log-$SLURM_JOBID-$SLURMD_NODENAME-$SLURM_LOCALID.log 2>&1

\end{lstlisting}

\begin{center}
  \textbf{Script 3.} keras-tuner-dyanamic.py
\end{center}
\lstset{language=python}
\begin{lstlisting}
def set_environment(num_gpus_per_node="8"): # 'num_gpus_per_node' should be set based on a HPC spec.
    import os
    nodename = os.environ['SLURMD_NODENAME']
    procid = os.environ['SLURM_LOCALID']
    print(nodename)
    print(procid)
    stream = os.popen('scontrol show hostname $SLURM_NODELIST')
    output = stream.read()
    oracle = output.split("\n")[0]
    print(oracle)
    if procid==num_gpus_per_node:
        os.environ["KERASTUNER_TUNER_ID"] = "chief"
        os.environ["CUDA_VISIBLE_DEVICES"] = "0"
    else:
        os.environ["KERASTUNER_TUNER_ID"] = "tuner-" + str(nodename) + "-" + str(procid)
        os.environ["CUDA_VISIBLE_DEVICES"] = procid
 
    os.environ["KERASTUNER_ORACLE_IP"] = oracle + ".ib.bridges2.psc.edu" # Use full hostname
    os.environ["KERASTUNER_ORACLE_PORT"] = "8000"
    print("KERASTUNER_TUNER_ID:    %s"%os.environ["KERASTUNER_TUNER_ID"])
    print("KERASTUNER_ORACLE_IP:   %s"%os.environ["KERASTUNER_ORACLE_IP"])
    print("KERASTUNER_ORACLE_PORT: %s"%os.environ["KERASTUNER_ORACLE_PORT"])
 
def main():
    # User's tuning code goes here, e.g.,
    # - Import necessary packages including Keras Tuner
    # - Read dataset
    # - Normalize/scale if necessary
    # - Define a 'hypermodel' (a search space is specified here)
    # - Define a 'tuner' (a search algorithm is specified here)
    # - Execute hyperparameter  searching
 
if __name__ == '__main__':
    set_environment()
    main()
 

\end{lstlisting}

\end{document}